\documentclass[11pt,a4paper]{article}
\usepackage[hyperref]{acl2021}
\usepackage{times}
\usepackage{graphicx}
\usepackage{latexsym}
\usepackage{algorithm}
\usepackage{flushend}

\usepackage[noend]{algpseudocode}

\usepackage{microtype}

\aclfinalcopy % Uncomment this line for the final submission
\title{Generative Conversational Networks}

\author{Alexandros Papangelis, 
  Karthik Gopalakrishnan,
  Aishwarya Padmakumar, \\
  {\bf Seokhwan Kim, 
  Gokhan Tur,
  Dilek Hakkani-Tur} \\
   Amazon Alexa AI \\
  }

\date{}

\begin{document}
\maketitle
\begin{abstract}
 Inspired by recent work in meta-learning and generative teaching networks, we propose a framework called Generative Conversational Networks, in which conversational agents learn to generate their own labelled training data (given some seed data) and then train themselves from that data to perform a given task. We use reinforcement learning to optimize the data generation process where the reward signal is the agent's performance on the task. The task can be any language-related task, from intent detection to full task-oriented conversations. In this work, we show that our approach is able to generalise from seed data and performs well in limited data and limited computation settings, with significant gains for intent detection and slot tagging across multiple datasets: ATIS, TOD, SNIPS, and Restaurants8k. We show an average improvement of 35\% in intent detection and 21\% in slot tagging over a baseline model trained from the seed data. We also conduct an analysis of the novelty of the generated data and provide generated examples for intent detection, slot tagging, and non-goal oriented conversations.
\end{abstract}

\section{Introduction}
In the past few years, large language models (some with tens of billions of parameters) have shown great success and have propelled the field of Natural Language Processing (NLP) and the industry forward. In parallel, recent advances in Meta Learning have shown great promise in computer vision, robotics, and machine learning in general (see \cite{hospedales2020meta} for a survey), as these approaches have the potential to overcome deep learning challenges such as data bottlenecks, computation requirements, and generalization. All of these challenges are particularly relevant to conversational AI, as we are still lacking large annotated conversational datasets, but we have orders of magnitude larger generic text data. Moreover, it can be very costly to annotate such data in their entirety and train high-performing task-specific conversational agents.

%While such models usually benefit from more data and more parameters, this is not always the case for conversational agents where oftentimes smaller, more focused data achieve better results (more in-domain utterances, less hallucinations, etc.). 
% Aishwarya: I think that statement that smaller more focused data is good for conversations needs a cite
% Nowadays we have large conversational datasets, some more annotated than others, and orders of magnitude larger generic text data. However, it can be very inefficient to use such data in their entirety to train high-performing task-specific conversational agents with Meta-Learning. 
% As the authors of \cite{such2020generative} note: \\
% \emph{“For example, recent work in curriculum learning (Graves et al., 2017), active learning (Konyushkova et al., 2017; Settles, 2010) and core-set selection (Sener et al., 2018; Tsang et al., 2005) demonstrates that {\bf a surrogate dataset can be created by intelligently sampling a subset of training data, and that such surrogates enable competitive test performance with less training effort}”.}

By adopting recent advances in Meta-Learning and Neural Architecture Search, we envision the next generation of intelligent conversational agents, that can create the data they need in order to train themselves to perform a task. We take a step towards this direction by adapting Generative Teaching Networks (GTNs) \cite{such2020generative} from image recognition (MNIST, CIFAR10) to conversational AI and training it with Reinforcement Learning (RL) using Proximal Policy Optimisation (PPO) \cite{ziegler2019fine}. Our approach, called \emph{Generative Conversational Networks} (GCN), allows a conversational agent to generate its own annotated training data and uses RL to optimize the data generation process. It then uses that data to train an agent to perform according to given specifications. These specifications can refer to any language-related task, from intent detection to full task-oriented conversations. 

\begin{figure*}[t]
    \centering
    \includegraphics[width=0.75\textwidth]{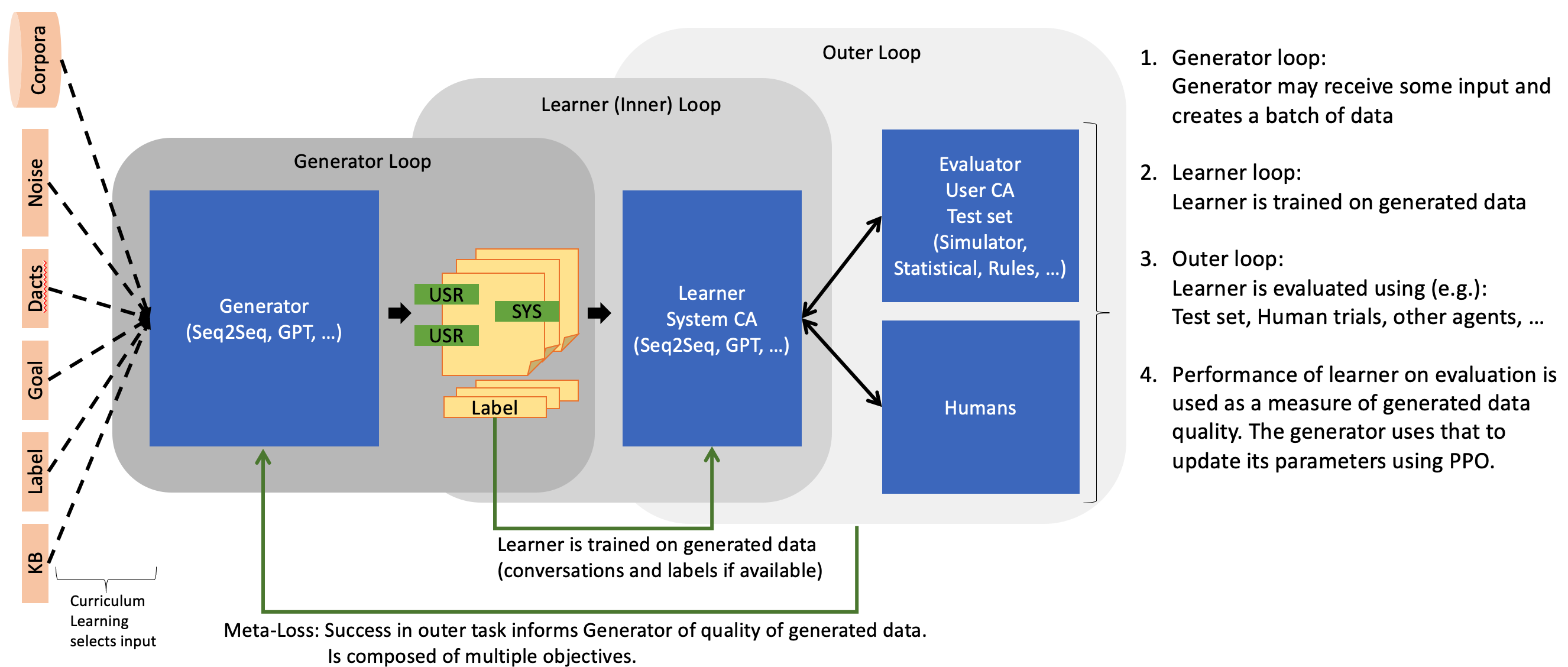}
    \caption{Generative Conversational Networks Architecture. We use PPO as described in \cite{ziegler2019fine} to perform the generator update using the meta-loss. USR refers to the user side and SYS to the system side.}
    \label{fig:arch}
    \vspace{-5mm}
\end{figure*} 

Similar to Generative Adversarial Networks (GAN), GCN effectively trains two models, a data generator and a learner. Unlike GAN-based approaches, however, GCN do not require a discriminator, only a numerical reward that can be obtained by any means and reflects the performance of the learner. This frees the architecture from tight domain constraints and allows it to be more adaptive and creative; some analysis and examples are shown in the respective section. Moreover, contrary to earlier approaches \cite[e.g.]{hou2020c2c}, we do not generate delexicalised utterances therefore we are not limiting our models to the vocabulary that exists in the data nor do we require a vocabulary to be provided. This allows GCN to better generalise from seed data, and create annotated training examples that are task-focused but also diverse and help increase the overall performance. 

Potential use cases for GCN include quick prototyping when limited resources are available, or when human feedback is available for training to continuously adapt to changes in the incoming data. GCN can also be applied when creating simulated agents with different characteristics (roles, personalities, etc) that can be used for training or evaluation. Our main contributions can be summarized as follows:
\begin{itemize}
\itemsep -0.5ex
    \item We propose GCN, a meta-learning approach for training conversational agents using RL
    \item We demonstrate that GCN can generalise from seed data in limited-resource settings (data and computation) and achieve competitive performance in two NLP tasks: intent detection and slot tagging
    \item We show that GCN can also be applied to multi-turn, non-goal oriented conversations.
\end{itemize}

\section{Related Work}
There have been plenty of prior works in few-shot learning for dialogue tasks including natural language understanding~\cite{shah2019robust,liu2020coach,hou2020few}, dialogue state tracking~\cite{wu2019transferable,dingliwal2021few} and response generation~\cite{tran2018adversarial,mi2019meta,chen2020few,peng2020few}, % 
which aim to make each model transferable to a low-resource new domain.
Another line of recent work proposes data augmentation techniques for conversational agents~\cite{campagna2020zero,kale2020few,lee2021neural}.
While these studies focus on one-time augmentation by heuristics or static neural models, our proposed approach keeps improving the data generation and hence models trained with that data, using RL. 

%Other approaches use model-based RL \cite{wu2019switch}, Multi-Agent RL \cite{georgila2014single,liu2017iterative,papangelis2019marl}, or GANs \cite{liu2018adversarial} for task-oriented interactions.

C2C-GenDA (cluster to cluster generation for data augmentation) \cite{hou2020c2c} is a generative data augmentation approach focused on slot filling. This method jointly encodes multiple realisations (i.e. a cluster) with the same semantic interpretation and generates multiple previously unseen realisations. A ``duplication-aware attention" model guarantees that there are no replications of the input in the output, since the model receives all realisations of a given semantic interpretation. The authors train their model with paraphrasing pairs and show that they outperform existing systems. Contrary to our work, C2C-GenDA generates delexicalised utterances that need to be post-processed. 

With SC-GPT \cite{peng2020data}, the authors finetune GPT-2 on dialogue act - utterance pairs on two scenarios, when the ontology is available (i.e. many valid dialogue act sequences are available) or when unlabeled data sets are available (i.e. many valid utterances are available). They finetune for each condition differently and achieve good results for intent and slot tagging. Our approach is different in that we directly generate annotated data and do not require large data for fine-tuning.

PROTODA \cite{kumar2021protoda} is a method similar in spirit to our work in that it uses seed data and generates new data to train intent classifiers. The authors use prototypical networks that are trained on a large number of intents and are evaluated on unseen intents, showing good performance. Our approach is more universal and geared towards multiple conversational AI tasks.

\section{Generative Conversational Networks}
Following \cite{such2020generative} and \cite{ziegler2019fine}, we propose a new Meta-Learning architecture combining the two, for training conversational agents using RL. Our approach can be helpful in settings with limited resources, or in settings where we want to augment data along some dimension (e.g. dialect, terminology, small talk, user types, expand to other domains, etc.).

\subsection{Generative Teaching Networks}
Generative Teaching Networks (GTNs) \cite{such2020generative} is a meta-learning approach to generate synthetic supervised data to train AI systems. Specifically, GTNs are data-generating networks that given Gaussian noise and a label in the input, generate data. The input label is optional as GTNs can also produce labelled data. This data is used by another model (e.g. a classifier) and the performance of the second model on a given task is then used as a loss signal to train the GTN. Eventually, GTNs learn to generate good quality data so that the classifier model can perform well on the given task. GTNs have been successfully applied to train MNIST \cite{MNIST} and CIFAR10 \cite{CIFAR10} classifiers from synthetic data with very good performance and, besides supervised tasks, they can be applied to unsupervised and reinforcement learning. A broader application of GTNs is to evaluate candidate neural architectures in neural architecture search.

\subsection{GCN Architecture}
We pair GTNs with \cite{ziegler2019fine}, who use PPO to train transformers from human feedback. \footnote{Using the Transformer Reinforcement Learning (TRL) implementation: https://github.com/lvwerra/trl}
Using RL to optimize the data generation process is crucial to generalize from the training data\footnote{Theoretically, we can train the generator from scratch using noise in the input. We have not tested this condition in this work, however.}, as we discuss later in the paper (section 5.4). We compute a reward for each datapoint rather than for each batch or for the entire generated data, to provide a more fine-grained signal which allows GCN to better handle the complexities of conversational tasks and avoid language degradation.

Figure \ref{fig:arch} shows an overview of the GCN architecture. It has three main parts: a) a data generator, b) a learner, and c) an evaluator. The training process iterates over the following steps until good performance is achieved: a) a generation step, where data is generated in batches; b) a learner training step, where a new learner model is spawned and trained on the data provided by the generator; and c) a generator update step, where the learner is evaluated on a validation set or by humans using the learner and feedback is provided back to the generator. Algorithm 1 describes the training process.

\begin{algorithm}
\caption{GCN training procedure.}
\begin{algorithmic}[1]

\Procedure{Train}{$D_{seed}$, $D_{val}$, $D_{test}$}
    \State Initialize Generator $g$
    \If{$D_{seed}$}         
        \State $g$.train($D_{seed}$)
    \EndIf

    \While{Performance$_{meta} < \epsilon$}   \Comment training
        \State $D_{gen} \leftarrow g$.generate()
        \State $D \leftarrow Curriculum(D_{gen}, D_{seed}$)
        \State Sample and initialize new Learner $l$ 
        \State $l$.train($D$)
        \State Performance$_{meta} \leftarrow l$.evaluate($D_{val}$)
        \State $g$.update(Performance$_{meta}$)
    \EndWhile  
    
    \State $D \leftarrow g$.generate()  \Comment evaluation
    \State Sample and initialize new Learner $l$
    \State $l$.train($D$)
    \State $l$.evaluate($D_{test}$)    \Comment{or other evaluator}
\EndProcedure
\label{alg:gcn}
\end{algorithmic}
\end{algorithm}

The generator can be any model of choice. It generates data on demand and can receive various kinds of input, depending on the configuration and task: noise to encourage diverse data, specific labels to generate focused data, goals, dialogue acts, or knowledge base results to encourage task-oriented dialogues, and so on. The generator's output will be a batch of data that is then sent to a learner model. At each meta-iteration, a new learner is created either from a pool of available model architectures or using the same type of model (our approach in this work). The learner is trained on the generated batches of data using a held-out validation set (generated or provided) and its performance on the validation set is used as a reward to train the generator using PPO. After the training phase, the generator trains a new, final learner that is evaluated on an external test set, never seen by the generator or any learner, or by a human or an evaluator agent. In theory, GCN can train the generator and the learner from scratch; in practice, however, we rely on pre-trained models for the generator and the learners, to speed up the process. We use a distilled version of GPT2 (distilGPT2, 82M parameters) to demonstrate the power of GCN without requiring very large models.

We implement a form of curriculum learning by providing the learner with seed data and gradually introducing generated samples. This is done at batch-level, to avoid cases where some batches contain mostly good examples and some contain mostly bad ones, in the early stages of training. As the training progresses, the percentage of generated data grows to 100\%. Other forms of curriculum learning are left for future work (i.e. one can provide the generator with labels from which to generate utterances, or goals, dialogue states, and knowledge base entries to generate dialogues, etc.). Equation 1 shows how we calculate the number of learner training iterations that contain seed data (warmup iterations $i_w$) at each meta-iteration $i_{meta}$ (data generation \& learner training cycle) and equation 2 shows how we calculate the number of datapoints ($n_{wb}$) per batch during the warmup iterations:

\begin{equation}
    i_{w} = \frac{I_{warmup} - i_{meta}}{I_{warmup}} I_{learner}
\end{equation}
where $i_{w}$ is the number of warmup learner iterations for the current meta-iteration $i_{meta}$. $I_{warmup}$ is the number of meta-iterations for which we have warmup learner iterations and $I_{learner}$ is the number of learner iterations at each meta-iteration.

\begin{equation}
    n_{wb} = \frac{|b_{gen}|}{I_{warmup}}(I_{warmup}-i_{meta})
\end{equation}
where $n_{wb}$ is the number of datapoints in the current learner iteration batch that will be pulled from the seed data (the rest are generated) and $|b_{gen}|$ is the generator's batch size.

\subsection{Data Generation}
Since our generator is a GPT-2 based model, we train it using special tokens that act as separators between labels and utterances: \\
$<$BOS$>$ label $<$GO$>$ utterance $<$EOS$>$

If we want the generator to create labelled data, we prompt it with a $<$BOS$>$ token (our approach in the experiments); if we want to provide the label and get a corresponding utterance, we prompt it with $<$BOS$>$ label $<$GO$>$. Depending on the task, the \emph{label} can be an intent, a collection of slot-value pairs, a previous utterance, etc.:
\begin{itemize}
\itemsep -0.5ex
    \item $<$BOS$>$ \emph{flight} $<$GO$>$...
    \item $<$BOS$>$ \emph{people 5 time after 9am} $<$GO$>$...
    \item $<$BOS$>$ \emph{previous utterance} $<$GO$>$...
\end{itemize}
%\\
for intent detection, slot tagging, and conversational response generation, respectively.
Each learner will receive data in this format and will have to parse it to retrieve the input (between $<$GO$>$ and $<$EOS$>$) and the target label (between $<$BOS$>$ and $<$GO$>$) in order to train itself. When training for the slot tagging task, we convert all slot names to words or phrases (e.g. convert ``arrival\_time" to ``arrival time") in the label portion of the input to better take advantage of distilGPT2. In this setting, the generator outputs IOB tags in addition to the output described previously and those tags are used as the learner's labels. 

For more complex tasks such as task-oriented dialogues, we can use more special token separators to separate the various kinds of input. Alternatively, we can design task-specific generators where GPT-2 can be a part of the model and we can have other encoders and decoders for the various kinds of optional inputs (belief states, goals, etc.).

\subsection{Learner Training}
{\bf Intent Detection.} For this task we use a RoBERTa-base sentence classifier \cite{roberta} as a learner. Upon receipt of a batch of data, the learner will parse it and create an \emph{input} and a \emph{target} tensor, containing the utterances and labels respectively.\\
{\bf Slot Tagging.} For this task we use a RoBERTa-base slot tagger \cite{roberta}. Similarly to intent detection, the learner will parse the batch of data but using the utterance part to create the \emph{input} tensor and the IOB tags to create the \emph{target} tensor. \\
{\bf Non-goal oriented interaction.} For this task we use the Bert2Bert model \cite{rothe2020leveraging} where, similarly to intent detection, the learner will create the \emph{input} and \emph{target} tensors that represent one dialogue turn. 

\subsection{Generator Training}
\label{sec:generator_training}
Following \cite{ziegler2019fine}, we use two generator models, $\pi$ and $\rho$. $\pi$ is the model that is being trained and $\rho$ is a reference model (distilGPT2 in our case) that keeps $\pi$ from diverging too much, via a Kullback-Leibler (KL) term in the reward function. PPO is then used to update $\pi$.

In GCN, each datapoint created by the generator is saved as is the performance of the learner for that particular datapoint. When the generator is being trained, we combine the per-datapoint performance $P_{d}$ with the validation performance $P_{meta}$ of the learner to compute the reward:
\begin{equation}
    R_d = \alpha P_{meta} + (1-\alpha) P_{d}
\end{equation}
\noindent where $d$ is the datapoint, $R_d$ is the reward for that datapoint, and $P$ is a measure of performance, e.g. accuracy, F1 score, perplexity, etc.. In our experiments, we use equal weighting for the reward components: $\alpha=0.5$. $R_d$ is then used to train the generator $\pi$:
\begin{equation}
    R(d, a) = R_d - \beta log \frac{\pi(a|d)}{\rho(a|d)}
\end{equation}
\noindent where $a$ is the {\it ``action"}, i.e. the system's response and the coefficient $\beta$ is varied dynamically (see \cite{ziegler2019fine} for details). After some pre-defined number of training epochs, we copy the parameters of $\rho$ to $\pi$.

\subsection{Training from Human Feedback}
One of the benefits of using RL to train GCN is that it allows for continuous adaptation based on human feedback. In a GCN-trained production system, for example, we can combine human ratings with other metrics (appropriateness, time lag, factual correctness, etc) to compute a reward signal. As the rated conversations include the human side as well, that reward can only be used to characterise the batch of GCN-produced data that were generated to train the agent in production. Using reward shaping methods \cite[e.g.]{el2013reward, su2015reward}, we can derive a reward per individual conversation or even per dialogue turn. 

\section{Experiments}
We assess GCN along two dimensions, creativity in data generation and task performance. Regarding task performance, we conduct experiments in limited-resource settings along two tasks across four datasets and compare against baseline models. Specifically, we conduct few-shot experiments where for each experiment we allow a limited number of updates (100 learner iterations for the learners and 15 meta-iterations for the generators). We use a batch size of 10 for intent detection and 50 for slot tagging. We evaluate GCN on the following tasks:

% \begin{figure}[t]
%     \centering
%     \includegraphics[width=0.45\textwidth]{LearningCurve.png}
%     \caption{GCN+RL performance on TOD intent classification VS inner iterations.}
%     \label{fig:learning_crve}
% \end{figure}

{\bf Intent detection.} For intent detection, similarly to \cite{kumar2021protoda}, we evaluate our approach on Facebook's Task-Oriented Dialogues (TOD) \cite{schuster2019cross}, ATIS \cite{hemphill1990atis}, and SNIPS \cite{coucke2018snips} using random samples of the data of various sizes (from 0.5\% to 10\%). In this setting, the generator produces pairs of utterances and intent labels. The learner is a RoBERTa-base sentence classifier.

{\bf Slot tagging.} For slot tagging we use TOD, SNIPS, and the Restaurants8k dataset \cite{coope2020span}, again using random samples of the data of various sizes (from 0.5\% to 10\%). In this case, the generator produces slot-value pairs and utterances that realise them exactly. The learner is a RoBERTa-base token classifier. In these initial experiments, we generate the tags via approximate matching, by looking at the label (slots and values) produced by the generator and finding them in the utterance that is also produced by the generator. Since we ask the generator to produce a structured dataset, we found that if we also ask it to produce IOB tags (i.e. asking the generator to learn how to do tagging) the system became very fragile due to small misalignments that result in low rewards.

\begin{table}[t]
\centering
\begin{tabular}{||c c c c||}
 \hline
 \multicolumn{4}{|c|}{\bf Baselines} \\
 \hline
 \multicolumn{4}{|c|}{Intent Classification (Accuracy)} \\
 \hline
 ATIS & TOD & SNIPS & \\ [0.5ex] 
 \hline
 0.929 & 0.963 & 0.939 & \\
 \hline
 \multicolumn{4}{|c|}{Slot Tagging (F1 Score)} \\
 \hline TOD & Restaurants8k & SNIPS &\\
 \hline
 0.969 & 0.92 & 0.938 & \\
 \hline \hline
 \multicolumn{4}{|c|}{\bf GCN+RL} \\
 \hline
 \multicolumn{4}{|c|}{Intent Classification (Accuracy)} \\
 \hline
 ATIS & TOD & SNIPS & \\ [0.5ex] 
 \hline
 0.956 & 0.99 & 0.944 & \\
 \hline
 \multicolumn{4}{|c|}{Slot Tagging (F1 Score)} \\
 \hline TOD & Restaurants8k & SNIPS &\\
 \hline
 0.968 & 0.947 & 0.943 & \\
 \hline
\end{tabular}
\caption{Performance at 5000 training iterations.}
\label{tab:baselines}
\end{table}

\begin{table}[t]
\centering
\begin{small}
\begin{tabular}{||c|c c c c c||} 
 \hline
 \multicolumn{6}{||c||}{ATIS Accuracy (100 learner iterations)} \\
 \hline
 & 0.5\% & 1\% & 2\% & 5\% & 10\%  \\
 \hline
 Base & 0.532 & 0.516 & 0.72 & 0.695 & 0.78 \\  
 GCN-RL & {\bf 0.738} & {\bf 0.757} & 0.769 & 0.78 & 0.803 \\
 GCN+RL & 0.732 & 0.734 & {\bf 0.809} & {\bf 0.816} & {\bf 0.851} \\
 \hline\hline
 
 \multicolumn{6}{||c||}{SNIPS Accuracy (100 learner iterations)} \\
 \hline
 & 0.5\% & 1\% & 2\% & 5\% & 10\% \\ [0.5ex] 
 \hline
 Base & 0.262 & 0.292 & 0.344 & 0.661 & 0.686 \\  
 GCN-RL & 0.229 & 0.424 & 0.547 & 0.715 & 0.783 \\
 GCN+RL & {\bf 0.602} & {\bf 0.638} & {\bf 0.734} & {\bf 0.798} & {\bf 0.865} \\[1ex]
 \hline \hline
 
 \multicolumn{6}{||c||}{TOD Accuracy (100 learner iterations)} \\
 \hline
 & 0.5\% & 1\% & 2\% & 5\% & 10\% \\ 
 \hline
 Base & 0.7 & 0.706 & 0.71 & 0.765 & 0.769 \\
 GCN-RL & 0.78 & 0.855 & 0.84 & 0.904 & 0.899 \\
 GCN+RL & {\bf 0.836} & {\bf 0.895} & {\bf 0.903} & {\bf 0.927} & {\bf 0.959} \\
 \hline

\end{tabular}
\end{small}
\caption{Intent detection limited-resource results various random subsets of the data.}
\label{tab:intents}
\end{table}

\begin{table}[t]
\centering
\begin{tabular}{||c|c||} 
 \hline
 \multicolumn{2}{|c||}{SNIPS-3 } \\
  \hline\hline
  PROTODA & 0.881  \\  
  GCN-RL & 0.822 \\
  GCN+RL & {\bf 0.926} \\[1ex]
 \hline
\end{tabular}
\caption{Results on the SNIPS-3 test set. We allow 5000 learner iterations here for a fairer comparison.}
\label{tab:protoda}
\end{table}

\subsection{Experimental Setup} 
We use the original train / validation / test splits provided with each dataset. For Restaurants8k, we randomly split the training set into training (80\%) and validation (20\%). Specifically for ATIS, we remove intents with less than 20 utterances as per \cite{kumar2021protoda}. To conduct our limited-resource experiments, we sample the respective percentage of training and validation data, making sure we preserve the distribution of classes as much as possible\footnote{We make sure that there is at least one datapoint for each intent / slot.} and always evaluate on the full test set. We pre-train the generator with the available training data of each few-shot setting and use a curriculum batch schedule to mix seed and generated data. The learner is trained on those batches for 100 iterations and once the iterations are finished, the learner is evaluated on the sampled validation set and its performance is used as a reward for training the generator. After 15 meta-iterations, the generator creates a final dataset that is used to train a learner that is evaluated on the held-out test set. To show the value of training the generator with RL, we compare two conditions against the baselines: \emph{GCN-RL}, where the generator used to augment the data is finetuned with the seed data but not trained with RL (this can be thought of as ``GTN for text" instead of image recognition), and \emph{GCN+RL} where the generator is finetuned and trained using RL. 

\begin{table}[t]
\begin{small}
\centering
\begin{tabular}{||c | c c c c ||} 
 \hline
 \multicolumn{5}{|c|}{SNIPS Intent classification (accuracy)} \\
 \hline
 & 1\% & 2.5\% & 5\% & 10\%\\ [0.5ex] 
 \hline
 C2C-GenDA & 0.481 & - & 0.679 & -\\ 
 (encoder-decoder) & & & &\\
 \hline
 SC-GPT & - & {\bf 0.941} & - & {\bf 0.981} \\
 (GPT-2) & & & &\\
 \hline
 GCN-RL & 0.907 & 0.901 & 0.906 & 0.926 \\
 (distilGPT2) & & & &\\
 \hline
 GCN+RL & {\bf 0.914} & 0.917 & {\bf 0.934} & 0.939 \\
 (distilGPT2) & & & &\\
 \hline
 \end{tabular}
\caption{Comparison with C2C \cite{hou2020c2c} and SC-GPT \cite{peng2020data} on few-shot intent detection. We allow our learners to train for 5000 iterations.}
\label{tab:c2c}
\end{small}
\end{table}

\subsection{Training Details}
Training a GPT-2 model with PPO in the context of GCN can be sensitive to hyperparameters for a variety of reasons, the most important being that we receive a numerical reward that characterises an entire batch of data. As mentioned in section \ref{sec:generator_training}, calculating per-datapoint performance seems to help speed up training. An option we do not explore in this work is to calculate per-token rewards. We also find that if we gradually unfreeze the generator's layers during training, the training becomes more stable. These strategies make training fairly stable and robust to hyperparameter values and apart from setting an appropriate learning rate, no other hyperparameter tuning was needed. We use the following PPO hyperparameters ($lr$: learning rate):
\begin{itemize}
\itemsep -0.5ex
    \item $\beta = 0.2$ (adaptive)
    \item train for 4 epochs per batch
    \item $lr_{generator} = $1e-5
    \item $lr_{learner} = $3e-3 (intents)
    \item $lr_{learner} = $1e-4 (slots)
    \item $lr_{learner} = $1e-4 (chit-chat)
\end{itemize}

We train the learners using Adam \cite{kingma2014adam} and we train the generator using Stochastic Gradient Descent because we found it to be much more stable than Adam.

\section{Task Results}
\label{sec:task_results}
In this section, we present the results of our evaluation; all reported numbers are averages of 3 runs. We conduct limited-resource experiments, i.e. restricting the available computation as well as the available data. We show that we achieve an average improvement of 35\% in intent detection and 21\% in slot tagging over a baseline model trained from the seed data. 

As the focus of our work is on a novel training framework, we do not explicitly compare against few-shot approaches (that would take the place of the learner model) and typically do not restrict computation. However, for completeness, we compare against approaches that are similar to ours and not specifically designed for one task.

\subsection{Baselines}
We use the learners trained directly on the available seed data as our baselines. Table \ref{tab:baselines} shows the performance of our learners (Baselines) when trained directly on each dataset for 5000 iterations using all available training data and the performance of GCN+RL under the same conditions.

\subsection{Intent Detection} 
Table \ref{tab:intents} shows the limited-resource experiments where we compare GCN to the baseline (RoBERTa sentence classifier). \emph{Base} refers to the baseline, \emph{GCN-RL} refers to GCN without RL fine-tuning, and \emph{GCN+RL} refers to GCN with RL finetuning. We see that GCN+RL outperforms the other conditions in all settings. 

In Table \ref{tab:protoda}, we show a comparison with PROTODA \cite{kumar2021protoda} in the SNIPS-3 setting. In that setting, the evaluation is performed on 3 intents: \emph{GetWeather}, \emph{PlayMusic}, and \emph{SearchCreativeWork}, and training is performed on ATIS, TOD, and SNIPS.

In Table \ref{tab:c2c}, we show a comparison with C2C-GenDA \cite{hou2020c2c} and SC-GPT \cite{peng2020data} on SNIPS. GCN outperforms C2C-GenDA while SC-GPT performs better than GCN, which is expected since it is based on GPT-2 (instead of distilGPT2) and fine-tuned on 400K additional dialogue act - utterance pairs. Another reason may be that we allow 5000 learner iterations for GCN due to computation resource constraints which could explain the lower performance.

\subsection{Slot Tagging}
Table \ref{tab:slots} shows the results from our limited-resource experiments for slot tagging. Similarly to the previous task, we see that \emph{GCN+RL} outperforms the other conditions in most settings but we do see less gains here compared to \emph{GCN-RL}. This can be explained by the increased complexity of the data the generator is required to produce: slots, values, and corresponding utterances (compared, for example, to intents and corresponding utterances). Such complexity means that small mistakes (generating paraphrases of slots or values, over or under generation of the corresponding utterance, other misalignments) can cause the learner to under perform and thus lead to that datapoint receiving a very low reward, even though only a small mistake occurred. In future work, we are looking to alleviate this by working with per-token rewards.

\section{Non-Goal-Oriented Interactions} 
To demonstrate the ability of GCN to handle conversational tasks, we use TopicalChat \cite{Gopalakrishnan2019} and train a Bert2Bert learner. The generator here produces utterance pairs if prompted with the $<$BOS$>$ token, or produces a response if prompted with $<$BOS$>$utterance$<$GO$>$. To produce a batch of data, we first prompt the generator with a $<$BOS$>$ token and observe its output pair ($u$, $u^\prime$). For the next turns, we prompt the generator with $<$BOS$>u^\prime<$GO$>$, observe its output $u^{\prime\prime}$, and feed that to the following turn. Table \ref{tab:tc_example} shows example data generated by GCN that do not exist in the TopicalChat dataset. We leave a thorough evaluation for future work.

\begin{table}[t]
\centering
\begin{small}
\begin{tabular}{||c | c c c c c||} 
 \hline
 \multicolumn{6}{|c|}{TOD F1 (100 learner iterations)} \\
 \hline
 & 0.5\% & 1\% & 2\% & 5\% & 10\%\\ [0.5ex] 
 \hline
 Base & 0.541 & 0.567 & 0.617 & 0.723 & 0.741 \\ 
 GCN-RL & 0.558 & 0.689 & 0.793 & 0.748 & 0.86 \\
 GCN+RL & {\bf 0.597} & {\bf 0.728} & {\bf 0.815} & {\bf 0.838} & {\bf 0.868} \\
  \hline\hline
 \multicolumn{6}{||c||}{Restaurants8k F1 (100 learner iterations)} \\
 \hline
 & 0.5\% & 1\% & 2\% & 5\% & 10\% \\ [0.5ex] 
 \hline
 Base & 0.182 & 0.36 & 0.627 & 0.626 & 0.774 \\ 
 GCN-RL & 0.313 & 0.481 & 0.633 & 0.622 & 0.771 \\
 GCN+RL & {\bf 0.334} & {\bf 0.564} & {\bf 0.659} & {\bf 0.696} & {\bf 0.827} \\
 \hline\hline
  \multicolumn{6}{||c||}{SNIPS F1 (100 learner iterations)} \\
 \hline
 & 0.5\% & 1\% & 2\% & 5\% & 10\% \\ [0.5ex] 
 \hline
 Base & {\bf 0.347} & 0.454 & 0.618 & 0.705 & 0.77 \\ 
 GCN-RL & 0.342 & {\bf 0.494} & 0.654 & 0.782 & 0.819 \\
 GCN+RL & 0.326 & 0.483 & {\bf 0.719} & {\bf 0.804} & {\bf 0.899} \\
 \hline
 
%   \hline\hline
%   \multicolumn{6}{|c|}{CONLL F1 (100 learner iterations)} \\
%  \hline
%  & 0.5\% & 1\% & 2\% & 5\% & 10\% \\ [0.5ex] 
%  \hline
%  Base & 0.255 & 0.26 & 0.393 & 0.437 & 0.44 \\ 
%  GCN-RL & 0.32 & 0.285 & 0.432 & 0.478 & 0.52 \\
%  GCN+RL & {\bf 0.335} & {\bf 0.358} & {\bf 0.468} & {\bf 0.51} & {\bf 0.556}\\ [1ex] 
%  \hline
\end{tabular}
\end{small}
\caption{Slot tagging limited-resource F1 results.}
\label{tab:slots}
\end{table}

\begin{table*}[t]
\centering
\begin{small}
\begin{tabular}{||c | c c | c c | c c ||}
 \hline
 \multicolumn{1}{||c|}{} & \multicolumn{2}{|c|}{Seed EM} & \multicolumn{2}{|c|}{Train EM} & \multicolumn{2}{|c||}{Self-BLEU} \\
 \hline
 ATIS \% & GCN-RL & GCN+RL & GCN-RL & GCN+RL & GCN-RL & GCN+RL \\
 \hline
 0.5\% & 1.57\% & 0.0\% & 0.0\% & 17.45\% & 0.977 & 0.982\\
 1\% & 0.37\% & 0.0\% & 0.0\% & 5.82\% & 0.996 & 0.971\\
 2\% & 0.37\% & 0.23\% & 0.63\% & 7.72\% & 0.997 & 0.974\\
 5\% & 3.27\% & 0.68\% & 0.3\% & 8.34\% & 0.998 & 0.967\\
 10\% & 7.83\% & 1.08\% & 1.0\% & 6.6\% & 0.997 & 0.966\\
 100\% & 66.33\% & 15.97\% & 14.33\% & 15.97\% & 0.985 & 0.963\\
 \hline
\end{tabular}
\end{small}
\caption{GCN exact match (EM) wrt the seed or the full train data and Self-BLEU scores on ATIS (micro avg).}
\label{tab:creativity}
\end{table*}

\begin{table*}[t]
\centering
\begin{small}
\begin{tabular}{||c | c ||} 
 \hline
  Intent & Utterance \\ [0.5ex] 
  \hline
  flight+airfare & \$5 or less on the fly from boston to atlanta\\
  city & is there one way on i-town on august eighteenth\\
  flight & what continental flights leave phoenix on friday\\
  reminder set & i want to be reminded to finish seasoning the steaks\\
  \hline \hline
  {\bf Slots} \& Values & Utterance \\ [0.5ex] 
  \hline
  {\bf weather} jacket & do i need a light jacket today?\\
   {\bf datetime} today & \\
  \hline
  {\bf datetime} for the first  & set an alarm for the first of\\
  of every month & every month for flea and tick prevent\\
  \hline
  {\bf generic} & cancel my earliest alarm \\
  \hline
  {\bf object\_type} tv series  & look for the tv series   \\
  {\bf object} all around & all around \\
  performance horse & performance horse weekly\\
  weekly & \\
  \hline
  {\bf movie} the fox and the fox & what time does the fox play\\
%   \hline
% \end{tabular}
% \end{small}
% \caption{A mix of good and bad examples generated by GCN. The errors may be at the label or utterance part.}
% \label{tab:intent_slot_examples}
% \vspace*{-2mm}
% \end{table*}

% \begin{table*}[t]
% \centering
% \begin{small}
% \begin{tabular}{||c | c ||} 
 \hline\hline
 Speaker & Utterance \\ [0.5ex] 
 \hline
 SP 1 & Hi, how are you today?  \\
 SP 2 & I'm great! how are you? \\
 SP 1 & I am well, thanks! I am a fan of football. Are you?  \\
 SP 2 & A little, I know there is a league. Some players in the NFL are really competitive. \\
 SP 1 & Interesting. I used to watch it all the time, but I don't really watch a lot anymore. \\ 
 & I think it's sad they don't get a chance anymore.\\
 \hline
\end{tabular}
\end{small}
\caption{A mix of good and bad examples generated by GCN. The errors may be at the label or utterance part.}
\label{tab:tc_example}
\end{table*}

\begin{figure}[t]
    \centering
    \includegraphics[width=0.5\textwidth]{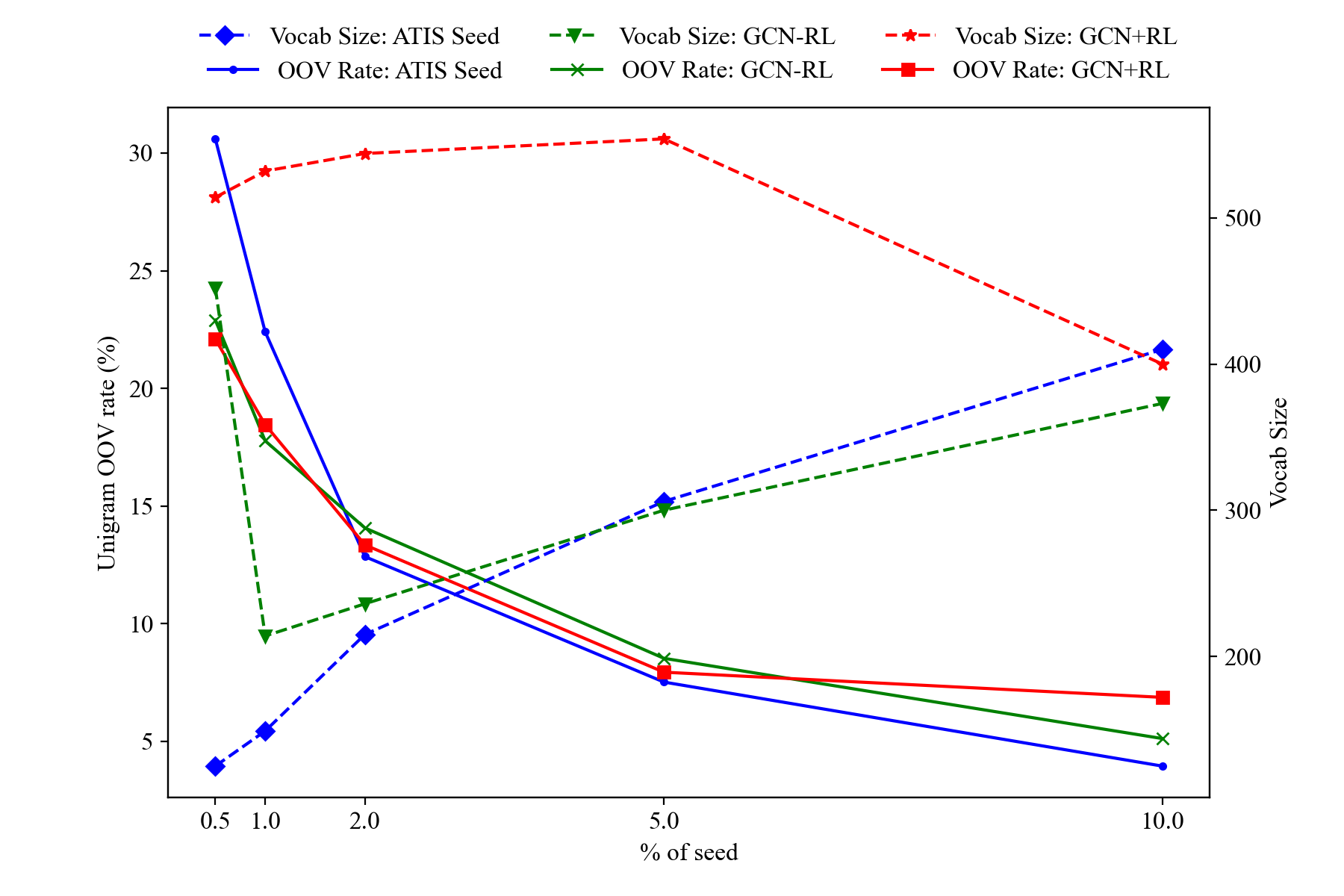}
    \caption{Unigram out of vocabulary rates and vocabulary sizes with respect to the ATIS test set.}
    \label{fig:oov}
\end{figure}

\section{GCN Generator Creativity}
To better understand the quality of the generated data, we analyze the \emph{creativity} of GCN, or how many examples are copied from the data vs created or paraphrased. We compare the seed data with data generated by GCN-RL and GCN+RL choosing ATIS as our use case. We calculate exact match rates (EM) with respect to the seed data and Self-BLEU scores \cite{zhu2018texygen} in Table \ref{tab:creativity} and unigram OOV rates (OOV) with respect to the test set and vocabulary sizes in Figure \ref{fig:oov}. We see that GCN-RL is more influenced by the seed data as the seed data size grows but when trained with RL it maintains a higher OOV rate. While not all OOV words are good, this trend in combination with the results on section \ref{sec:task_results} means that GCN creates more diverse data that are focused on the task and this is why we see the increase in task performance. As we can see from Table \ref{tab:creativity}, RL helps reduce repetitions in the data and GCN in general creates data outside of the seed but that are valid (a larger portion exist in the full train data).

This means that GCN learns to produce good quality novel data that can be used to train higher performing learners. It is clear from the results in section \ref{sec:task_results} that applying RL to GCN helps generate more diverse data, that in turn result in higher task performance. For instance, using 10\% of the data, after 15 meta-iterations, the data generated by GCN+RL achieve an average 94.4\% of the top baseline performance (Table \ref{tab:baselines}) using 2\% of the training iterations on intent detection. For slot tagging, we achieve an average of 91.8\% of the baseline performance.

Table \ref{tab:tc_example} show some example datapoints generated by GCN+RL in all three tasks.

\section{Conclusion}
We have presented \emph{Generative Conversational Networks}, an approach that takes a step towards conversational agents that generate their own data and learn to perform well in conversational tasks. We conducted an analysis on GCN's creative ability and demonstrated its performance and efficiency on two sample language understanding tasks, intent detection and slot tagging.
However, GCN has the potential to perform many more tasks and we are currently evaluating it for non-knowledge- and knowledge-grounded conversations. As future work, we will investigate per-token rewards as well as having populations of learners with different architectures evaluated on the same task, and having learners evaluated on multiple tasks.

\bibliographystyle{acl_natbib}
\bibliography{acl2021}

%\appendix

\end{document}